\pdfoutput=1

\documentclass[11pt]{article}

\usepackage[preprint]{acl}

\usepackage{times}
\usepackage{latexsym}

\usepackage[T1]{fontenc}

\usepackage[utf8]{inputenc}

\usepackage{microtype}

\usepackage{inconsolata}

\usepackage{float}
\usepackage{todonotes}
\usepackage{amsmath}
\usepackage{amssymb}
\usepackage{comment}
\usepackage{bm}
\usepackage[inline]{enumitem}

\usepackage{fancyhdr}

\title{On the Limits of Multi-modal Meta-Learning with Auxiliary Task Modulation Using Conditional Batch Normalization}

 \author{
Jordi Armengol-Estapé{\thanks{Equal contribution.}}{\thanks{Work done while interning at Mila.}}$^1$ \quad
Vincent Michalski$^{\ast 2,3}$ \quad
Ramnath Kumar$^4$ \\
\textbf{Pierre-Luc St-Charles$^2$ \quad
Doina Precup$^{2,5,7}$ \quad
Samira Ebrahimi Kahou$^{2,6, 7}$} \\
$^1$University of Edinburgh \quad
$^2$Mila\quad
$^3$Université de Montréal\quad
$^4$Google Research \\
$^5$McGill University \quad
$^6$University of Calgary \quad
$^7$CIFAR \\
\texttt{jordi.armengol.estape@ed.ac.uk}\vspace{-4mm}
}

\begin{document}
\maketitle
\begin{abstract}
Few-shot learning aims to learn representations that can tackle novel tasks given a small number of examples. Recent studies show that cross-modal learning can improve representations for few-shot classification. More specifically, language is a rich modality that can be used to guide visual learning. In this work, we experiment with a multi-modal architecture for few-shot learning that consists of three components: a classifier, an auxiliary network, and a bridge network. While the classifier performs the main classification task, the auxiliary network learns to predict language representations from the same input, and the bridge network transforms high-level features of the auxiliary network into modulation parameters for layers of the few-shot classifier using conditional batch normalization. 
The bridge should encourage a form of lightweight semantic alignment between language and vision which could be useful for the classifier.
However, after evaluating the proposed approach on two popular few-shot classification benchmarks we find that a) the improvements do not reproduce across benchmarks, and b) when they do, the improvements are due to the additional compute and parameters introduced by the bridge network. We contribute insights and recommendations for future work in multi-modal meta-learning, especially when using language representations. 
\end{abstract}

\begin{figure}[t]
\includegraphics[width=0.45\textwidth]{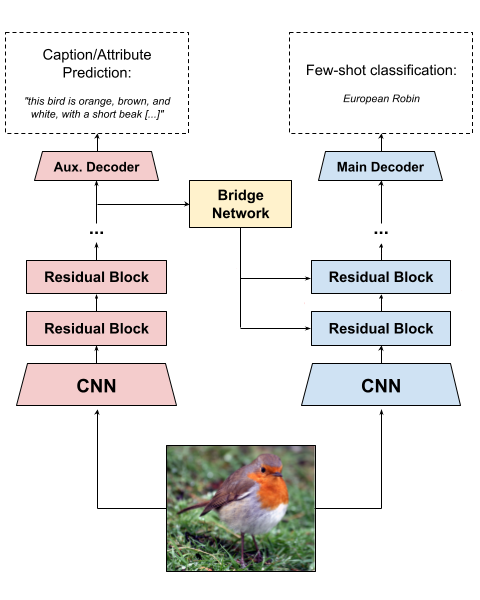}
\centering
\caption{Architectural overview of the method we experimented with. It consists of three components: a classifier, an auxiliary network, and a bridge network. The few-shot classifier and auxiliary network receive the same input example. The bridge network transforms high-level features of the auxiliary network into modulation parameters for layers of the few-shot classifier through conditional batch normalization. }
\label{fig:overview}
\end{figure}

\section{Introduction}

It is widely recognized that humans can learn new concepts based on very little supervision, i.e. with few examples (or ``shots''), and generalize these concepts to unseen data \citep{lake2011one}. Recent advances in deep learning on the other hand have mostly relied on datasets with large amounts of labeled examples, primarily due to overfitting concerns in low data regimes. Although the development of better data augmentation and regularization techniques can alleviate these concerns, many researchers now assume that future breakthroughs in low data regimes will emerge from either transferring generic models pretrained on very large datasets with unsupervised objectives \citep{devlin-etal-2019-bert,NEURIPS2020_1457c0d6}, or from meta-learning, i.e. ``learning-to-learn''. Here, we study the problem of learning-to-learn in few shots by using an embedding space in which we perform classification using a similarity metric. In this meta-learning setting, a model is trained on a handful of labeled examples at a time under the assumption that it will learn how to correctly project examples of different classes and generalize this knowledge to unseen labels at test time.

Although this setting is often used to illustrate the remaining gap between human capabilities and machine learning, we could argue that the lack of context poses a serious disadvantage to machine learning models. Indeed, these models typically work based on a single-pass analysis while humans can first look at and understand contextual information before trying to interpret new classes \citep{swingley2010fast}. It has been observed many times in the past that training models with contextual information such as auxiliary modalities can help build a more robust task-independent feature space \citep{ruder2017overview, elliott2016multimodal, radford2021learning}. Auxiliary tasks however often require large support datasets with good label distributions and a delicate adjustment of network capacity to really help improve performance on the main task \citep{alonso2016multitask}. Multi-modal information can be difficult to process using a simple backbone architecture due to the varied structure and high-level nature of some typically used modalities, although recent Transformer-based works have shown it's possible, albeit costly \citep{jaegle2021perceiver}. 
We refer to the Appendix \ref{sec:appendix-rel} for a more comprehensive study of the related work.


We propose studying whether multitask learning with multi-modal objectives could be beneficial for few-shot learning even with commonly-used low-capacity feature extraction backbones, and without weight sharing between the main and auxiliary tasks. We study a way to condition multiple layers of our main feature extractor using an embedding produced by an entirely separate auxiliary network working on the same input data. The conditioning is applied to normalization layer parameters using a bridge network and it helps specialize the representations produced by the main feature extractor without affecting its architecture. Our idea here is to mimic the way humans can leverage context to help solve the recognition problem by combining low-level and high-level cues. In other words, we allow the main feature extractor to decide ahead of time what it should focus on based on task-level contextual knowledge. The proposed model architecture is illustrated in Figure~\ref{fig:overview}. In contrast with previous works that also studied feature extraction conditioning and multi-modal learning, our approach is simple and can be applied to any feature extractor with batch normalization layers. The bridged-parallel-network design we propose also simplifies the feature alignment process since both branches process the same input data. Finally, the need for only a single input modality at test time leads to a more practical design for downstream applications. 

However, after evaluating the proposed approach on two popular few-shot classification benchmarks we find that a) the improvements do not reproduce across benchmarks, and b) when they do, the improvements are due to the additional compute and parameters introduced by the bridge network. We contribute insights and recommendations for future work in multi-modal meta-learning, especially when using language representations.

%

\section{Proposed method}

In this section, we formulate conditional batch normalization in the context of few-shot learning. We propose a model, SimpAux, with two feature extractors that predict high-level (language-based) attributes of images as well as their semantic class. The embeddings of the attribute prediction pipeline (or ``auxiliary'' pipeline) are used to condition the batch normalization layers of the main visual feature extractor, which is based on a ProtoNet architecture \citep{snell2017prototypical}. More specifically, we use ProtoNet++ improvement introduced in \citet{oreshkin2018tadam}, with a Resnet-12  \citep{resnet}, which is a common choice in few-shot learning settings \citep[e.g.][]{oreshkin2018tadam,jiang2018learning}. The conditioning happens through a bridge connection composed of dense layers that translates the auxiliary embedding into batch normalization statistics. These three components are shown in Figure~\ref{fig:overview} and are described in the following subsections. Note that we use the same input modality (imagery) for the auxiliary and main feature extractors. However, our method is not limited to this modality: it was primarily chosen for compatibility with existing datasets. We refer the reader to Appendix~\ref{app:background} for a review of the fundamental ideas required to better understand our proposed few-shot learning solution from a technical standpoint.

\subsection{Auxiliary visual processing}

The auxiliary network in our proposed approach is agnostic of the main network's architecture and task. To simplify comparisons with a wider number of few-shot learning methods and to improve practicality, we formulate this network as a second visual processing pipeline that converts the same images fed to the main network into different embeddings. The multi-modal nature of our overall design comes from the supervised task used to learn the auxiliary network's embeddings: its goal is to predict language-based information from the images. More specifically, we experimented with predicting a) Attributes, available in datasets such as CUB-200-2011~\cite{wah2011caltech}, with cross-entropy, soft F1, or multi-label soft margin loss functions, and b) caption embeddings, with cosine similarity loss on the sentence embeddings emitted by  SentenceBERT \cite{reimers-2019-sentence-bert}. We ended up using multi-label soft margin loss as it was simpler and the other approaches did not provide significant improvements. However, for the datasets for which attributes were not available, we resorted to the sentence embeddings approach. As for the auxiliary model architecture itself, we also use a ResNet-12 as we do for the classifier.

\subsection{Conditioning bridge}

The role of the conditioning bridge is to transform the embeddings generated by the auxiliary network into an array of $\bm{\gamma}$ and $\bm{\beta}$ parameters that can be used in the various batch normalization layers of the main network. In contrast with late representation fusion strategies, e.g. the one of \citet{de2017modulating}, this strategy allows for the early modulation of the main feature extraction pipeline with the high-level semantic information extracted from the auxiliary pipeline. Our hypothesis is that this information provides adequate context to dynamically adapt the main feature extractor while keeping its original architecture intact (and thus simple).

Since the distribution of the input representation varies at each layer of that network, the normalization parameters also need to be unique for each layer. We define our bridge as a multilayer perceptron (MLP) with a fixed intermediate representation size and an output size that corresponds to twice the total size of batch normalization layers in the main network (to account for both $\bm{\gamma}$ and $\bm{\beta}$). 

\section{Experimental results}

We evaluate SimpAux against the baseline, ProtoNet++, (the improved version of ProtoNets suggested in \citet{oreshkin2018tadam}) on two popular few-shot learning benchmarks, CUB-200-2011~\cite{wah2011caltech} and mini-ImageNet~\cite{vinyals2016matching} in 5-shot learning settings, using attributes for CUB and embeddings on synthetic captions for Mini-Imagenet for the auxiliary visual processing network. We refer to Appendix \ref{sec:appendix-impl} for additional implementation details.

Table \ref{tab:cub-results} shows the results of ProtoNet++ and SimpAux on CUB 5-shot. Our model clearly outperforms the baseline by a margin of around 1.5 points in accuracy.
\begin{table}[H]
\centering
\begin{tabular}{lr}
\hline \textbf{Model} & \textbf{Accuracy (\%)} \\ \hline
ProtoNet++ & $88.5 \pm 0.5$ \\
SimpAux & $90.0 \pm 0.7$ \\
\hline
\end{tabular}
\caption{\label{tab:cub-results} Accuracy on CUB. Each model was trained with five random seeds. Reported is the mean accuracy with 95\% confidence intervals on 600 randomly generated test episodes.}
\end{table}

These positive results on CUB showed the promise of the proposed approach. However, in the case of Mini-Imagenet 5-shot, in Table \ref{tab:miniimagenet-results} we can see the results of ProtoNet++ and SimpAux on Mini-Imagenet. In this case, the baseline slightly outperforms the proposed method, but recall that here we are using synthetic captions.
\begin{table}[H]
\centering
\begin{tabular}{lr}
\hline \textbf{Model} & \textbf{Accuracy (\%)} \\ \hline
ProtoNet++ & $75.4 \pm 0.4 $ \\
SimpAux & $74.9 \pm 0.1$ \\
\hline
\end{tabular}
\caption{\label{tab:miniimagenet-results} Accuracy on mini-ImageNet. Each model was trained with five random seeds. Reported is the mean accuracy with 95\% confidence intervals on 600 randomly generated test episodes.}
\end{table}

Finally, to test the hypothesis that the reason why our approach outperforms the baseline in CUB but not in ImageNet is the quality of the captions, we design an ablation study. We introduce a variation of SimpAux in which we use the exact same bridge network, but without input from the auxiliary network, to see whether the improvements are actually coming from the captions information or the additional compute and parameters from the bridge network. We find that there is no significant improvement over this variant when using the captions, suggesting that the improvement comes from the additional compute and the parameters provided by the bridge network.

\section{Discussion and recommendations}

From the experimental results, we conclude that a) the improvements provided by SimpAux do not reproduce across benchmarks, and b) when these improvements do indeed take place, they seem to be due to the additional compute and parameters provided by the bridge network.
We hypothesize three non-mutually exclusive reasons why image captioning as auxiliary task modulation via conditional batch normalization did not consistently improve the results: 
 1) a lack of quality of the image captions, attributes, or caption embeddings, 2) the limited impact of the conditional batch normalization approach, and 3) the difficulty of learning the auxiliary task. While improving the quality of captions, attributes and caption embeddings with better annotations or more powerful models could alleviate 1), the following recommendations and observations look at other aspects involved in this work.

\paragraph{Caution when evaluating systems with auxiliary multi-modal information.} Training models with contextual information such as auxiliary modalities have been shown to build a more robust task-independent feature space \citep{ruder2017overview, elliott2016multimodal, radford2021learning}. However, spurious improvements with multi-modal data are not new. For instance, \citet{elliott-2018-adversarial} empirically raises doubts about whether existing multi-modal translation systems, combining visual and textual data, actually make use of the visual information. Similarly, we have seen the other way around: it is perfectly possible to outperform a unimodal baseline with a multi-modal one without actually making use of the textual information; SimpAux's improvements in CUB were due to the additional parameters introduced by the bridge network. Thus, we recommend extra care when concluding that multi-modal information helps in a certain task, which is definitely possible but could be due to other factors.

\paragraph{Importance of implementation details.} We experimented with different activation functions, including ReLU \cite{DBLP:journals/corr/abs-1803-08375}, SELU \cite{NIPS2017_5d44ee6f}, and SiLU \cite{DBLP:journals/corr/HendrycksG16, DBLP:journals/corr/abs-1710-05941}. We found that SiLU consistently yielded slightly better results across benchmarks and settings. Ensuring that weight decay was not applied to bias parameters, which is not the default behavior in PyTorch \cite{NEURIPS2019_9015}, also turned out to be key to reproducing few-shot works originally implemented in Tensorflow \cite{tensorflow2015-whitepaper}.

\paragraph{Hyperparameter search.} 
In the hyperparameter search, we generally observed consistent results. However, we also observed a few outliers, which can be particularly extreme under certain settings in few-shot learning, and if used as empirical evidence, could totally change the conclusions. Thus, we reiterate the need for reporting averages and variances instead of the results of a single run, and also recommend caution at extracting certain conclusions when performing large-scale hyperparameter searches, as noted by \citet{DBLP:journals/corr/abs-2109-08203}.

\paragraph{Advantages of the proposed architecture.} Our network architecture decouples task-specific branches: its bridge acts as a gate that selects relevant hints from the auxiliary network to influence the classification network. It is simpler than previous works that also studied feature extraction conditioning and multi-modal learning, and by design it requires a single input modality at test time, which simplifies practical deployments. 
SimpAux's architectural considerations are orthogonal to other few-shot learning research lines, and could be combined with them. Thus, we believe that, despite the limited success in the meta-learning setting, these architectural advantages could be a source of inspiration for future work.

\paragraph{Language-informed representations and few-shot learning.} Without episodic learning, \citet{radford2021learning} showed that language-informed visual representations can be successfully learned with large-scale supervised contrastive pretraining. Their approach, CLIP,  obtains high-performance at zero-shot classification. Leveraging their pretrained encoders could be interesting in the context of bootstrapping episodic learning with auxiliary tasks. It would however be difficult to guarantee that the classes used in few-shot settings have not been observed by CLIP during pretraining.


\section{Conclusion}
In this work, we have studied a new multi-modal architecture for few-shot learning consisting of an image classifier, an auxiliary network trained with image captions, and a modulating network based on conditional batch normalization to connect the two. While initially promising, we have observed the limits of this approach and how these limits could inform future research. 
\bibliography{acl_latex}

\appendix

\section{Related Work}
\label{sec:appendix-rel}


\textbf{Network conditioning.} Normalization layers have been used many times in the past as a means to influence the behavior of deep feature extractors. For example, early works in arbitrary style transfer studied how modulating instance normalization parameters could align representations across styles that are not already known at run time \citep{huang2017arbitrary, ghiasi2017exploring}. The flexibility gained by this modulation strategy has been adopted to tackle many other problems where feature extractors must dynamically change their behavior at run-time. For example, \citet{de2017modulating} and \citet{perez2017film} use conditional normalization layers to manipulate feature extractors in a selective manner for visual question answering and reasoning tasks. In the few-shot learning literature, \citet{oreshkin2018tadam} apply a form of normalization conditioning for task-dynamic feature extraction. In their case, instances are first encoded with an ``unconditioned'' feature extractor, and the resulting embeddings are used to condition the same feature extractor in a subsequent pass. In contrast, we base our conditioning on auxiliary labels and formulate a single-pass inference process. We also do not impose any constraints on the architecture of the main or auxiliary networks, meaning one can be much smaller than the other if required by the limited size of the dataset.

Note that there are also alternative conditioning strategies for few-shot learning paradigms that do not involve normalization layers. For example, embeddings can be directly modulated by a second network stage that analyzes the contextual information from the task \citep{ye2020few, qiao2019transductive}. Popular feature extractor architectures can also be slightly modified by adding conditionally shifted neurons to adapt representations using context at prediction time \citep{munkhdalai2018rapid}. Alternatively, the entire parameter set of various convolutional layers inside the feature extractor can be inferred at prediction time using a parallel network \citep{bertinetto2016learning, bertinetto2018meta, zhao2018dynamic}. A recent approach has also been proposed by \citet{chen2022vision} to adapt large-scale multi-modal transformer-based backbones. The downside to these solutions is the dependency on large networks that must learn complex modulation operations from the task context, or the use of a memory bank on which an attention mechanism can operate. In contrast, normalization conditioning is a more lightweight approach that is easier to learn in small data regimes due to the reduced complexity of the modulation factors (i.e. the normalization statistics).


\textbf{Recent trends in few-shot learning.} There have been far too many strategies proposed to tackle few-shot learning for us to inventory them here. For a survey and a modern taxonomy, we refer the reader to the work of \citet{wang2020generalizing}. Instead, we note that many researchers over the years have highlighted the lack of a universal evaluation methodology for these methods. Recent independent efforts have shown that many ``state-of-the-art'' solutions are actually quite fragile and can be outperformed by simple baselines when evaluated and compared properly \citep{chen2019closer, dhillon2019baseline, tian2020rethinking}. All of these works found that simple CNN backbones trained using a cross entropy loss and then optionally fine-tuned on test time queries can deliver competitive performance with respect to recent models. Transductive learning using test time queries in particular has been recently re-explored as an effective solution for few-shot learning \citep{dhillon2019baseline, ziko2020laplacian}. Such findings highlight that more research effort should be spent on model-agnostic robustness improvements and less on the introduction or tuning of new model architectures as well as their training regimes. Our work falls in line with this idea while also promoting the use of multi-modal labels for improved few-shot learning.

As for multi-modal few-shot learning itself: it is not a new approach to the problem, but it is also not a popular one, as typical benchmarks only focus on using only imagery as input. Nonetheless, multiple strategies have been proposed to help deal with data scarcity in few-shot learning. For example, \citet{pahde2019self} feed image captions to a generative model during training to obtain additional images of the target classes. Their method however relies on several pre-trained and notably hard-to-train model components. \citet{xing2019adaptive} and \citet{schwartz2019baby} also leverage caption data but instead combine visual and semantic representations to improve class discrimination in metric space. In contrast to our work, they rely on parallel feature extraction pipelines that are combined in a ``late fusion'' fashion, whereas we propose a way to modulate the entirety of any visual pipeline architecture with semantic information. \citet{vuorio2019multimodal} applies a similar modulation idea to the model-agnostic, meta-learning (MAML) framework of \citet{finn2017model}. In their case, they rely on the modulation layers proposed by \citet{perez2017film} to condition their main task network. \citet{tseng2020cross} follow the same strategy to deal with domain generalization issues in few-shot learning. In comparison, our proposed auxiliary network is trained in a supervised cross-modal setting where its embeddings are used to modulate our main network. Also, since we apply modulation through batch normalization, our approach can handle data samples that do not possess auxiliary labels or captions.

\section{Background}
\label{app:background}

Here, we review some of the fundamental ideas required to understand our proposed few-shot learning solution.

\subsection{Episodic few-shot learning and ProtoNets}

In episodic few-shot learning, an ``episode'' is represented as an $N$-way, $K$-shot classification problem where $N$ is the number of examples per class and $K$ the number of unique class labels. During training, the data in each episode is provided as a support set $S = \{(\bm{x}_{1,1},\bm{y}_{1}), ..., (\bm{x}_{N,K},\bm{y}_{N})  \}$ where $\bm{x}_{i, j} \in \mathbb{R}^D$ is the i-th instance of the j-th class, and $\bm{y}_{j} \in \{0, 1\}^K$ is its corresponding one-hot labeling vector. The goal in each episode is to optimize a function $f$ that classifies new instances provided through a ``query'' set $Q$ which contains instances of the same classes as $S$. This task is difficult because $N$ is typically very small (e.g. 1 to 10), the classes change every episode, and the actual test set used to evaluate a model does not contain classes that were seen in support sets during training.

We build our solution on top of Prototypical Networks \citep[ProtoNets; ][]{snell2017prototypical}, as it is now accepted as a good yet simple baseline. According to \cite{chen2019closer}, it is more robust than other recent few-shot learning approaches and it generalizes well across various dataset domains. ProtoNets tackle few-shot learning by learning an embedding space where each class is represented by a cluster, or \emph{prototype}. A prototype $\bm{c_k} \in \mathbb{R}^M$ for a class $k$ is simply defined as the mean of the instance embeddings that belong to $k$, that is:
\begin{equation}
    \bm{c_k} = \frac{1}{S_k} \sum_{(\bm{x}_{i,j},\bm{y}_{i}) \in S_k} f(\bm{x}_{i,j}),
\end{equation}
\noindent where $S_k$ is the support subset of all instances that belong to class $k$, and $f$ is a learned function. Next, the probability of assigning a new instance $x$ to a class $k$ is computed via the softmax of the distance to all class prototypes:
\begin{equation}
    p(y=k | \bm{x}) = \frac{\exp{-d(f(\bm{x}), \bm{c}_k)}}{\sum_{k'}\exp{-d(f(\bm{x}), \bm{c}_{k'})}},
\end{equation}
for any given distance function $d: \mathbb{R}^M \times \mathbb{R}^M \mapsto [0, +\infty)$.

\subsection{Batch normalization and conditioning}

Batch normalization was proposed by \cite{ioffe2015batch} as a solution to speed up training by reducing the problem of coordinating weight updates across the different layers of a model. In short, batch normalization performs a reparameterization on the intermediate representations of a model so that assumptions regarding their spread and distribution in subsequent layers will be less affected by stochastic updates. More specifically, given a batch of $n$ feature maps $B = \{ \bm{z}_1, ..., \bm{z}_n\}$ with $C$ channels each, batch normalization performs channel-wise reparameterization using
\begin{equation}
    \text{BN}(\bm{z}_{l,c} | B, \bm{\gamma}, \bm{\beta}) = \gamma_c \cdot \frac{\bm{z}_{l,c} - \mu_c}{\sigma^2_c + \epsilon} + \beta_c,
\end{equation}
\noindent where $\bm{\gamma}$ and $\bm{\beta}$ are vectors of learned channel-wise parameters, $\epsilon$ is a constant used for numerical stability, and $\mu_c$ and $\sigma^2_c$ are the mean and variation values computed across batch and spatial dimensions of $B$.

Many researchers now recognize that batch normalization has beneficial side-effects on the landscape of the optimization problem \citep{goodfellow2016deep, santurkar2018does}. These benefits have lead to the rapid adoption of this technique across the majority of new and popular model architectures. Consequently, the important role and ubiquitous nature of batch normalization make it an interesting target for the conditioning of models using auxiliary data. This idea was first introduced by \citet{de2017modulating}: they inject visual concepts from natural language in a visual processing pipeline for VQA by manipulating batch normalization parameters. These parameters are influenced by the embeddings produced with a recurrent network. One advantage of this approach is that it can help learn how to dynamically specialize a model at test time without drastically increasing its overall number of learnable parameters. This advantage is very interesting in the context of few-shot learning where only small datasets prone to overfitting are considered. 

\section{Other implementation details}
\label{sec:appendix-impl}

Our ProtoNet backbone is the improved version of the original method (coined ProtoNet++) suggested by \citet{oreshkin2018tadam} that includes residual connections between convolution layers (Resnet-12). 
 We implement the models and data loaders with PyTorch \cite{NEURIPS2019_9015} and Torchmeta \cite{deleu2019torchmeta}, a meta-learning library. We experimented with different activation functions, and SiLU \cite{DBLP:journals/corr/abs-1710-05941} yielded the best results.

In our experiments, we use the CUB-200-2011~\cite{wah2011caltech} and mini-ImageNet~\cite{vinyals2016matching} datasets. For CUB, we use the split of \citet{chen2019closer} and also experiment with the captions collected by \citet{reed2016learning}. For mini-ImageNet, we use the setting proposed by~\citet{ravi2016optimization}, with synthetic captions generated using an open-source implementation of a Transformer \cite{vaswani2017attention} for image captioning.\footnote{\url{https://github.com/saahiluppal/catr}}

Our implementation is publicly available on Github.\footnote{\url{https://github.com/jordiae/simpaux-release}}

\end{document}